# Mapping Languages:
# The Corpus of Global Language Use

Jonathan Dunn

Department of Linguistics
University of Canterbury

jonathan.dunn@canterbury.ac.nz
www.jdunn.name

*Abstract*

This paper describes a web-based corpus of global language use with a focus on how this corpus can be used for data-driven language mapping. First, the corpus provides a representation of where national varieties of major languages are used (e.g., English, Arabic, Russian) together with consistently collected data for each variety. Second, the paper evaluates a language identification model that supports more local languages with smaller sample sizes than alternative off-the-shelf models. Improved language identification is essential for moving beyond majority languages. Given the focus on language mapping, the paper analyzes how well this digital language data represents actual populations by (i) systematically comparing the corpus with demographic ground-truth data and (ii) triangulating the corpus with an alternate Twitter-based dataset. In total, the corpus contains 423 billion words representing 148 languages (with over 1 million words from each language) and 158 countries (again with over 1 million words from each country), all distilled from Common Crawl web data. The main contribution of this paper, in addition to describing this publicly-available corpus, is to provide a comprehensive analysis of the relationship between two sources of digital data (the web and Twitter) as well as their connection to underlying populations.

# 1 Gathering Global Language Data

This paper describes a corpus of global language use that is drawn from web-crawled data and systematically compared with both Twitter data and census-based demographic data. The purpose is to both (i) represent regional varieties of languages using a consistent collection methodology and (ii) provide a data-driven resource for understanding what languages are used where. As shown by the web-as-corpus paradigm (Baroni, et al., 2009; Majlĭs & Zabokrtsḱy, 2012; Goldhahn, et al., 2012; Benko, 2014), raw web data contains observations of language use that can be leveraged to create linguistic corpora. Further, these web-based corpora have been shown to represent local language use (Davies & Fuchs, 2015; Cook & Brinton, 2017) and can be compared with Twitter-based corpora which have themselves been shown to represent local language use (Grieve, et al., 2019). The Corpus of Global Language Use (CGLU: now at version 4.2)[1] sifts through data from 147 billion web pages in order to distill a corpus of approximately 423 billion words representing 148 languages and 158 countries with at least 1 million words each. This includes 1,916 language-country sub-corpora with at least 1 million words and 68 sub-corpora with at least 1 billion words. While previous iterations of this corpus have been used in existing work (Dunn, 2019a, 2019b; Dunn & Adams, 2019), the contribution of this paper is (i) to evaluate this expanded version of the corpus (increased from 16.6 billion to 423 billion words) and (ii) to analyze the degree to which digital data sources can be used to represent local language use.

This paper is structured as follows: First, we discuss how the web data was collected, cleaned, and organized. Second, the language identification model is motivated and evaluated on independent data to ensure that it performs well on diverse datasets. Third, a comparison corpus of 8 billion words from geo-located Tweets is described. Fourth, the web and Twitter corpora are evaluated against demographic data to understand the relationship between populations and digital language use. Fifth, we compare the web and Twitter corpora using standard corpus similarity measures in order to understand systematic differences between these sources of digital language data.

The goal of the CGLU is to systematically gather comparable language samples from every country in the world. The expectation is that some languages (e.g., Swahili) will be found only in certain regions of the world. Other languages (e.g., English and French) will be found in all regions and, as a result of their geographic distribution, will participate more widely in different language mixing situations. For the purposes of this paper, countries are grouped into sixteen larger geographic regions to simplify the analysis of language distribution. The distribution of the corpus across regions by number of words and by percentage of words is shown in Table 1. This table includes a previous iteration of the corpus (CGLU v.3), the currently described and greatly expanded version of the corpus (CGLU v.4.2), and the Twitter baseline corpus. The inventory of regions is relatively straight-forward. It is worth noting, however, that Brazil and Russia are large enough and produce enough language data that they are separated from surrounding countries.

The number of words for a given region depends on more than simply the population of the region: (i) the number of sites indexed by the Common Crawl; (ii) the population's degree of access to internet technologies; (iii) data cleaning decisions for this project that are subject to future improvements (i.e., identifying words across different writing systems). Although the relationship between words in the corpus and individuals in the regions is imperfect, in the aggregate this dataset can still be used to infer many things about language use around the world.

---

[1] The dataset is visualized at www.earthLings.io

*Table 1. Words Per Region*

|  | CGLU v.3 | | CGLU v.4.2 | | Twitter | |
|---|---|---|---|---|---|---|
| **Region** | **Words (mil)** | **%** | **Words (mil)** | **%** | **Words (mil)** | **%** |
| Africa, North | 123,859 | 0.74% | 1,223,532 | 0.29% | 203,867 | 2.53% |
| Africa, Southern | 59,075 | 0.35% | 26,868 | 0.01% | 159,807 | 1.99% |
| Africa, Sub | 424,753 | 2.55% | 5,938,870 | 1.39% | 571,644 | 7.10% |
| America, Brazil | 218,119 | 1.31% | 2,265,386 | 0.53% | 156,705 | 1.95% |
| America, Central | 886,610 | 5.32% | 8,877,634 | 2.08% | 852,793 | 10.60% |
| America, North | 236,590 | 1.42% | 51,921,657 | 12.15% | 452,263 | 5.62% |
| America, South | 1,163,008 | 6.98% | 22,441,384 | 5.25% | 824,502 | 10.25% |
| Asia, Central | 965,090 | 5.79% | 17,069,517 | 4.00% | 220,106 | 2.74% |
| Asia, East | 2,201,863 | 13.22% | 49,521,933 | 11.59% | 198,177 | 2.46% |
| Asia, South | 448,237 | 2.69% | 15,147,872 | 3.55% | 580,221 | 7.21% |
| Asia, Southeast | 2,011,067 | 12.07% | 21,386,781 | 5.01% | 443,258 | 5.51% |
| Europe, East | 4,553,101 | 27.34% | 65,413,609 | 15.31% | 748,654 | 9.30% |
| Europe, Russia | 101,444 | 0.61% | 15,363,644 | 3.60% | 135,778 | 1.69% |
| Europe, West | 2,422,855 | 14.55% | 143,748,386 | 33.65% | 1,703,436 | 21.17% |
| Middle East | 660,732 | 3.97% | 1,721,856 | 0.40% | 421,926 | 5.24% |
| Oceania | 164,025 | 0.98% | 1,743,571 | 0.41% | 372,623 | 4.63% |
| **TOTAL** | **16 billion** | **100%** | **423 billion** | **100%** | **8 billion** | **100%** |

One of the goals of the updated corpus is to achieve better coverage for under-represented areas such as South Asia, Sub-Saharan Africa, and Oceania. These areas have been severely under-represented in previous work, leading to systematically imbalanced datasets (c.f., Dunn & Adams, 2019). This much expanded corpus, on the one hand, provides significantly more language data from each of these areas. For instance, language-specific corpora for Hindi and Urdu have increased from 27 million to 586 million words and from 10 million to 112 million words, respectively. This enables many applications that the previous corpus could not support. On the other hand, the over-representation of certain areas has grown worse. For example, Western Europe accounts for only 5.7% of the world's population. But it accounts for 14.5% of CGLU v.3 and 21.1% of the Twitter baseline. The expanded CGLU v.4.2 has increased this over-representation to 33.6% of the corpus. In other words, improved methods for gathering the corpus have partly exaggerated the underlying bias of web data. This is not a problem in and of itself, however, for two reasons: First, population-based sampling could be used to create a geographically balanced sub-set of the corpus that does not over-represent western Europe. Second, a corpus of this size enables the representation of immigrant languages within Europe that supports new directions in corpus-based research. For example, the corpus now contains 18 million words of Turkish from Germany and 11 million words of Arabic from France. Although western Europe as a whole has a greater over-representation, this larger corpus enables the representation of minority populations within Europe. Given the goal of representing actual language use from populations around the world, this availability of geographic non-majority languages is an important achievement.

## 2 Processing Raw Web Data

This section presents the decisions made for processing the raw web data. For reproducibility, all code is provided in a public repository[2]. Language samples are geo-located using country-specific top-level domains: we assume that a sample from a web-site under the ".ca" domain is from Canada. This approach does not assume that whoever produced that sample was born in Canada or represents a traditional Canadian dialect group. Some countries are not available because their top-level domains are used for non-geographic purposes (i.e., ".ai", ".fm", ".io", ".ly", ".ag", ".tv"). Domains that do not contain geographic information are also removed from consideration (e.g., ".com" sites). An important improvement in CGLU v.4.2 is the inclusion of geographic TLDs that are not in a Latin script; this significantly increases the amount of data from languages like Hindi, Urdu, and Chinese that is collected. A complete list of TLDs is contained in the codebase.

The raw portions of the Common Crawl dataset used to build the corpus are shown in Table 2. The corpus uses every portion of the crawl from March 2014 to June 2019, totaling 147 billion web pages in total. No temporal divisions are included in the corpus because these dates represent the time of collection rather than the time of production: web data does not expire and there is a long-tail in which the same samples are observed multiple times across different periods. Deduplication can remove this long-tail but cannot add accurate time information.

*Table 2. Common Crawl Raw Data Size*

| Year | Period Represented (Months) | Pages |
|---|---|---|
| 2014 | March to December (8) | 22.53 billion |
| 2015 | January to December (10) | 17.98 billion |
| 2016 | January to December (9) | 16.91 billion |
| 2017 | January to December (12) | 37.28 billion |
| 2018 | January to December (12) | 36.30 billion |
| 2019 | January to June (6) | 16.05 billion |
| **Total** | **64 months** | **147.05 billion** |

The biggest challenge with web-crawled data is noise and duplication: we are after text that represents unique linguistic utterances, not lists or boilerplate or navigation words. For our purposes, a *sample* is any block of text that occurs within a <p> tag. Samples are discarded for a number of reasons: First, samples must reach a certain number of words. Second, samples cannot contain multiple instances of words related to "error," which often indicate an error page rather than actual content. Third, samples cannot contain more than four characters such as "|" that often

---

[2] https://github.com/jonathandunn/common_crawl_corpus

represent navigational structures. The complete set of heuristic selection rules can be found in the codebase provided. A key improvement for CGLU v.4.2 is the use of character identification in order to support a different length threshold for non-alphabetic scripts. In previous versions, languages like Chinese and Japanese were under-represented because of a naïve single length threshold.

The key technology for cleaning web-crawled data is deduplication: First, any samples that occur more than once on a single website are removed immediately. The idea is that many static portions of a website (copyright notices, slogans, navigation menus) can be removed simply because we are only interested in unique content. This deduplication takes a single web site as its scope. Second, the crawl is processed chronologically. This allows us to remove duplicate text that occurs across more than one site in a single month. The idea is that many text samples, for example a widely shared article, will appear multiple times in a single crawl. These sorts of texts are not unique and, more importantly, do not necessarily represent language use from a specific location. Thus, any sample that occurs more than once in a single time period is removed from the dataset. This methodology allows the dataset to be cleaned within feasible scopes: within the scope of a single website and within the scope of a single crawl. In both cases deduplication is performed at the level of the <p> tag, meaning that a larger web page may have some of its parts removed while others are retained in the corpus.

One of the challenges of a large multi-lingual corpus is that languages differ in the appearance of words. Character segmentation is used for Chinese (with the Jieba package[3]) and for Japanese (with the TinySegmenter package[4]). Although character segmentation is not used for other non-alphabetic languages, a character detection package is used to reduce the length threshold for non-alphabetic languages. Symbols, urls, hash-tags, at-mentions, and emojis are removed before the word limit is enforced; thus, samples must meet the length threshold after cleaning.

## 3  Evaluating Language Identification

An important part of preparing a global language use dataset is to reliably identify as many languages as possible (a task referred to as *LID* for "language identification"). This section presents an evaluation of the LID component (called idNet) using data independent of the web corpora and the Twitter baseline corpus. LID performance can be measured in terms of (i) the number of languages covered, (ii) the number of domains or registers covered, (iii) the sample size required, and (iv) overall prediction accuracy (c.f., Baldwin & Liu, 2010). The goal here is to maximize these measures while still eliminating the reliance on platform-specific training data (i.e., without using annotated Twitter training data). The end result is that idNet achieves an F1 above 0.95 for 464 languages in 50-character samples.

The dataset used for training and evaluating idNet contains several independent sources of data, shown in Table 3. A *sample* in Table 3 is a sequence of 50 characters, the window size that is used for language identification. Some of these data sources are considered ground-truth (e.g., Bible translations) while others are boot-strapped data used only for training purposes (e.g., Web2Corpus data). Only results for ground-truth data sources are shown in Table 3. The reason for the 50 character sample size is that (i) anything shorter has severely reduced accuracy but (ii) anything longer makes it difficult to work with the Twitter baseline corpus.

---

[3] https://www.github.com/fxsjy/jieba
[4] https://pypi.org/project/tinysegmenter

The first set of registers comes from a traditional LID source: religious texts. Bibles are taken from Christodoulopoulos & Steedman (2015) and from Brown (2014); Qurans are taken from the Tanzil corpus (Tiedemann, 2012). The second set contains official government and legislative texts: the European parliament, the JRC-Acquis corpus of European Union texts, and the United Nations (all from Tiedemann, 2012). The third set contains non-official formal texts: the EU Bookshop corpus (Skadiņš, et al., 2014), newspapers and commentary from GlobalVoices, NewsCommentary, and Setimes (all from Tiedemann, 2012), and Wikipedia (Majlĭs & Zabokrtsḱy, 2012). Moving to less formal registers, the fourth set contains documentation from open source software packages: Ubuntu and Gnome (from Tiedemann, 2012). The fifth set mimics conversational or informal speech: OpenSubtitles covering movies and television, TED Talks (both from Tiedemann, 2012), and Tatoeba for language-learning sentences (from tatoeba.org). The sixth set contains corpora representing specific languages: the Emille corpus of Indian languages (Baker, et al., 2004; Beta Release), the Indian Parallel Corpus (Post, et al., 2012), and the IARPA Babel project language packs (c.f., Andrus, et al., 2016). The seventh set contains data bootstrapped by applying existing LID models to web-crawled data: from Aranea (Benko, 2014), WaCky (Baroni, et al., 2009), and Web2Corpus (Majlĭs & Zabokrtsḱy, 2012). These last datasets are used for training but not for evaluation.

A Multi-Layer Perceptron classifier is used, with an architecture containing three dense layers (each with 300 neurons) with drop-out applied to each layer (0.25). Relu activations are used and predictions are made using a softmax layer. The feature set contains character trigram frequencies projected into a hashing space with 216k dimensions. Each training epoch is provided with an equal number of observations from each language-domain pair (1,000 samples). This allows less frequently observed languages to be modeled with the same accuracy as majority languages. It also avoids biasing the model toward language-domain pairs with a large number of samples. For example, many of the datasets contain millions of samples of English; without this sampling method, languages such as English or French would be over-represented and minority languages forgotten. While previous work (Liu & Baldwin, 2011) has viewed domain-independence as a feature selection problem (i.e., finding those features which are domain-independent), we view it as a sampling problem (i.e., ensuring that a high-capacity model learns the idiosyncrasies of each domain by supplying a large set of domains). This is important because the LID model needs to be able to perform well on short samples from a variety of registers while also including as many minority languages as possible.

The boot-strapped datasets rely on previously trained LID models (in this case, each dataset relies on a different model). Following previous work (Scannell, 2007), we use an additional existing LID tool (langid.py; Lui & Baldwin, 2012) to search these boot-strapped samples for contaminating majority languages: the text as a whole may belong to language A but a given 50-character sequence may not. Samples predicted to contain English, Spanish, or French are removed. This is important to keep minority languages from being contaminated by material in majority languages. This boot-strapping adds approximately 75 million training samples, which makes it possible to train a high-capacity classifier. Note that the performance of the model is evaluated only on ground-truth datasets to ensure that the boot-strapped training data does not influence the final evaluation. The dataset is divided into three functions: training data (shown to the classifier), testing data (used to evaluate the classifier during training epochs), and evaluation data (used for evaluating the final model, as shown in Table 3). The codebase for training the LID component is available[5], as is the final model used for processing the web corpus.[6]

*Table 3. LID Performance by Data Source with Off-the-Shelf Baselines (F1)*

| Domain | N. Langs | N. Test | idNet | CLD2 | langdetect | langid.py |
|---|---|---|---|---|---|---|
| Bibles | 85 | 76,611 | **0.98** | 0.56 | 0.45 | 0.54 |
| LTI (Bibles + UN) | 428 | 421,165 | **0.98** | 0.08 | 0.04 | 0.06 |
| Tanzil | 38 | 38,201 | **1.00** | 0.78 | 0.67 | 0.72 |
| Europarl | 21 | 21,109 | **1.00** | 0.97 | 0.99 | 0.99 |
| United Nations | 6 | 6,060 | **1.00** | 0.93 | 0.81 | 0.96 |
| JRC | 21 | 21,008 | **0.96** | 0.91 | 0.94 | 0.94 |
| EU Books | 25 | 24,442 | **0.98** | 0.93 | 0.91 | 0.95 |
| Global Voices | 31 | 28,565 | **0.98** | 0.88 | 0.82 | 0.91 |
| News Commentary | 10 | 10,075 | **1.00** | 0.95 | 0.88 | 0.99 |
| Wikipedia | 87 | 87,431 | **0.95** | 0.66 | 0.44 | 0.66 |
| Setimes | 8 | 8,080 | **1.00** | 0.84 | 0.86 | 0.84 |
| Gnome | 74 | 72,257 | **0.96** | 0.79 | 0.54 | 0.86 |
| Ubuntu | 71 | 66,673 | **0.96** | 0.80 | 0.61 | 0.86 |
| Open Subtitles | 45 | 45,450 | **0.98** | 0.90 | 0.76 | 0.89 |
| Tatoeba | 37 | 34,834 | **0.99** | 0.83 | 0.81 | 0.87 |
| TED Talks | 52 | 49,472 | **0.99** | 0.88 | 0.71 | 0.91 |
| IARPA Babel | 11 | 11,016 | **0.99** | 0.79 | 0.50 | 0.88 |
| Emille | 6 | 6,060 | **0.83** | 0.80 | 0.81 | 0.80 |
| Indian Parallel | 6 | 6,060 | **1.00** | 1.00 | 1.00 | 0.97 |
| Twitter (Over50)* | 25 | 23,791 | **0.96** | 0.96 | 0.96 | 0.93 |

The model covers 464 languages; importantly, this is evaluated across diverse registers at a small sample size. First, how stable is performance across domains? Table 3 shows F1 scores across ground-truth domains. The number of languages represented in each domain is also shown as a measure of the overall difficulty of language identification within that dataset. This shows that idNet achieves an overall F1 above 0.95 for all domains except Emille.

Why train a new LID component rather than using an existing off-the-shelf model? We compare idNet with three alternatives on the evaluation dataset: langid.py (Liu & Baldwin, 2012), CLD2 (Google, 2013), and langdetect (Google, 2014). These results clearly show that, in terms of language and domain coverage, existing models are not sufficient for building a global corpus. In situations where there are short samples from many registers, these models are not particularly accurate. On the one hand, this is not an entirely valid comparison because these off-the-shelf models are not trained to identify all of the languages present. However, even if we restrict ourselves to domains like the UN texts, EU Books, and Setimes (which only include languages within their purview), these off-the-shelf models are still not as accurate as idNet at this sample size. The point of this comparison is only to show that, for the purposes of organizing this corpus, a LID model needs to be robust across both languages and domains. It remains outside the scope of this paper to retrain each of the off-the-shelf models in order to see if, given different training conditions, they would achieve a more competitive accuracy. The point, rather, is to justify the introduction of a new LID component, idNet, which is more suited to the needs of this corpus-building project.

---

[5] https://github.com/jonathandunn/idNet

[6] https://publicdata.canterbury.ac.nz/Research/NZILBB/jonathandunn/idNet_models/

Why do we report F1 (a weighted combination of precision and recall) rather than simple accuracy? It is important to look at both false positives and false negatives because a geographic-centered crawl for language data will encounter many different and possibly unknown languages. As shown by the lower F1 scores of off-the-shelf models, a naïve approach to language identification here would skew the results by forcing predictions about unseen languages and unseen registers. The goal, then, is to provide an evaluation that is as diverse and as robust as possible.

A major reason for enforcing the 50 character sample size is to enable the comparison with the baseline Twitter corpus. No Twitter training data was used in preparing the idNet model. The official Twitter LID data (Twitter, 2015), containing 70 languages, is used for the evaluation (note that not all samples from the original dataset were still available when it was pulled for this study). Only those samples containing at least 50 characters (after cleaning) are included in the evaluation in Table 3. While idNet achieves an F1 of 0.96, this is not a better performance than the off-the-shelf models. This evaluation, however, shows that idNet can also be used on the comparison Twitter corpus so long as only Tweets containing at least 50 characters are included.

## 4 Collection and Preparation of Twitter Data

In isolation, web-crawled data provides a single observation of digital language use. Another common source of data is from Twitter (e.g., Eisenstein, et al., 2010; Roller, et al., 2012; Kondor, et al., 2013; Mocanu, et al., 2013; Eisenstein, et al., 2014; Graham, et al., 2014; Donoso & Sanchez, 2017). This paper uses a baseline Twitter corpus as a point of comparison: does the Common Crawl agree with Twitter data? We use a spatial search to collect Tweets from within a 50km radius of 10k cities taken from the GeoNames project.[7] This search method avoids biasing the selection of languages by relying on language-specific keywords or hashtags. Deduplication and text cleaning are used as described above for the main web-crawled corpus. Because the language identification component only has reliable predictions for samples with at least 50 characters (c.f., Section 3), a threshold of 50 characters is enforced after cleaning has taken place. The break-down of this cleaned comparison corpus by region is shown in Table 1 in Section 1; this represents two years of collection (July 2017 to July 2019).

## 5 Demographic Evaluation of Digital Corpora

The goal of representing local language use at a global-scale is only valid to the degree that these digital datasets (the web and Twitter) represent actual local populations.[8] In other words, we know that digital language data is biased by factors like per capita GDP and degree of internet access, so that poorer and less connected areas are likely to be under-represented. To this end we use ground-truth census-based demographic estimates to understand the biases of the corpus: the UN country-level population estimates (United Nations, 2017b), per capita GDP estimates (United Nations, 2017a), and country-level internet-usage statistics (United Nations, 2011).

---

[7] https://www.geonames.org

[8] The analysis presented in this section is also visualized in an open-source manner at https://www.earthlings.io

Starting with the density of the CGLU v.4.2 by number of words per country, Figure 1 shows that much of the corpus comes from North America, Western Europe, and Eastern Europe (countries in grey have no data, as a result of TLDs being excluded). On the other hand, the size of the corpus itself distorts this visualization: the US, Canada, France, and Spain all have more than 20 billion words each. Brazil, on the other hand, is under-represented by comparison but still has 2.2 billion words. This is contrasted in Figure 2 with the baseline Twitter corpus. First, fewer countries are missing because the Twitter corpus does not depend on TLDs for geo-referencing. Second, the Twitter corpus is more evenly distributed in South America and Southeast Asia. The scale is much smaller, though, with the US represented by 280 million words, Canada by 168 million, and Brazil by 156 million.

**Figure 1. CGLU v.4.2 by Words Per Country**

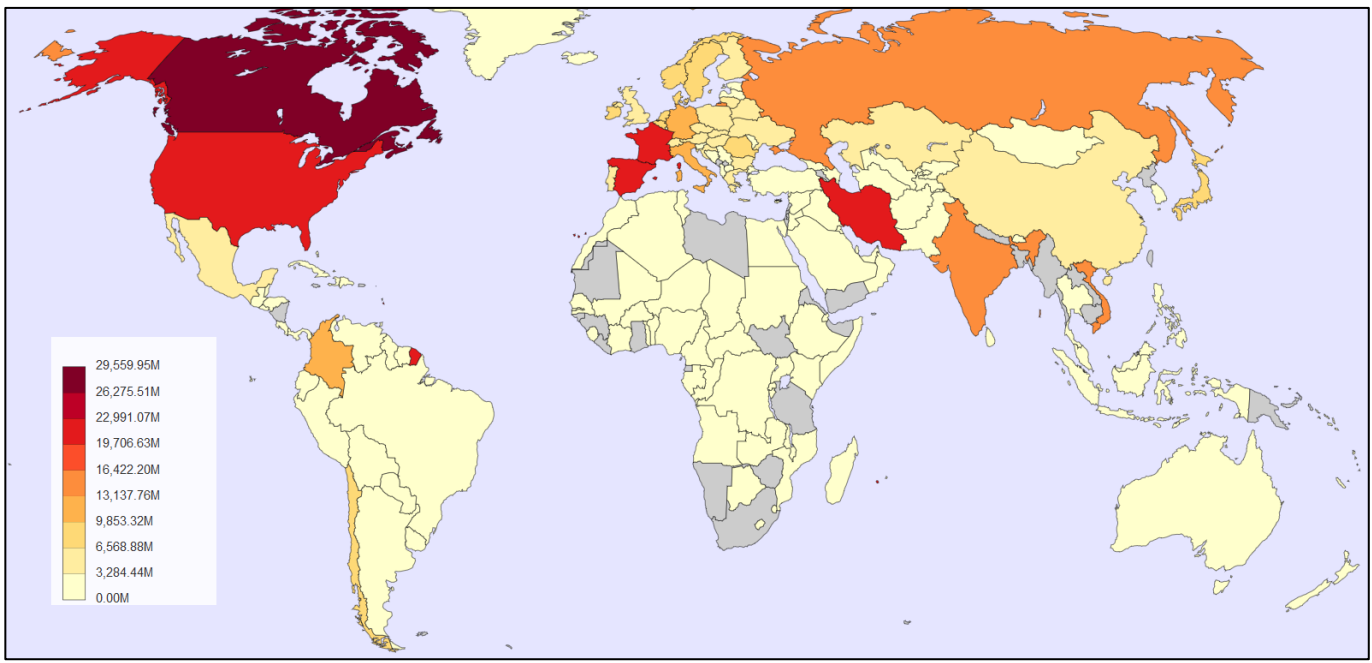

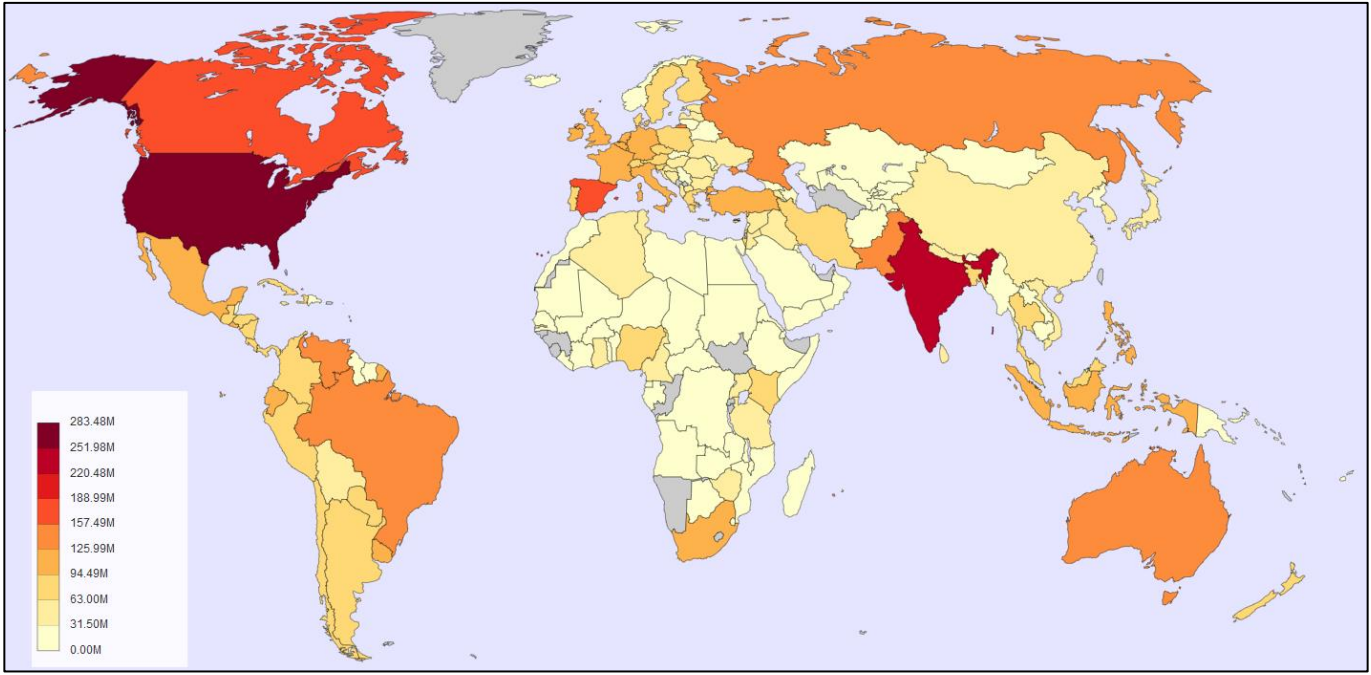

**Figure 2. Twitter Baseline Corpus by Words Per Country**

How closely do the three corpora relate (CGLU v.3, CGLU v.4.2, and Twitter) in terms of the number of words per country? First, CGLU v.3 and Twitter have a Pearson correlation of r=0.022, calculated over 198 countries with null values removed. CGLU v.3 and v.4.2 have a correlation of 0.294 for number of words per country. Why is the relationship between the geographic distributions of the corpora so low? First, CGLU v.4.2 has improved collection methods which increase the representation of both non-Latin urls and non-alphabetic scripts. Second, CGLU v.4.2 includes significantly more web pages in general (almost twice as many). Third, both CGLU v.3 and v.4.2 have ceilings on the number of web pages per country (otherwise the corpus would be too large to work with). As discussed in reference to Western Europe in Table 1, CGLU v.4.2 has a higher ceiling which means that some data-rich areas are more over-represented than in previous iterations of the corpus. Although the two web-based corpora have a low correlation between their geographic distributions, CGLU v.4.2 increases the correlation with the baseline Twitter corpus to 0.554 (v.4.2) from 0.022 (v.3).

But how well do either corpora represent ground-truth population density? The Twitter corpus has a correlation of 0.337 with estimated by-country population, compared with 0.372 (v.3) and 0.431 (v.4.2). This means that all three datasets weakly represent actual population distributions but that CGLU v.4.2 has achieved the best relationship. For example, the correlation between per capita GDP and words per country is 0.247 for Twitter and 0.223 for CGLU v.4.2, but only 0.090 for CGLU v.3. This means that the updated corpus better represents populations with higher GDP per person (increasing the wealth bias). The country-level estimates of access to internet technologies tells the same story: there is a correlation of 0.337 between internet access and the Twitter corpus size, similar to the 0.387 correlation for CGLU v.4.2. We see a trade-off here between (i) increasing the size of the corpus in general in order to have more language samples from under-represented areas and (ii) gathering an increasing number of samples from areas that are already over-represented. The advantage of having a larger corpus, however, is that it can be down-sampled to match population demographics if that is important for a particular application.

A more accurate model of a country's expected digital language production starts by adjusting raw population by the country's estimated internet access. In other words, if a country has a population of 100 million but only 50% have internet access, the *digital population* is only 50 million. We also expect wealthier populations to produce more language data, so we further weight this digital population measure by per capita GDP. This combined estimate has a correlation of 0.471 with Twitter and 0.573 with CGLU v.4.2. Thus, the best estimate of the density of digital language data (the number of words per country) is a combination of (i) population size, (ii) the population's access to the internet, and (iii) the population's per capita GDP. This means that, while the CGLU is not a perfect picture of local language use, it is a large and publicly-available corpus whose demographic biases we are able to quantify.

We have so far focused on density (number of words or number of people per country) without looking more closely at the linguistic properties of populations. Regardless of how much data there is per country, we can also quantify the relative language composition of that data. For example, the percentage of English use in CGLU v.4.2 is shown in Figure 3, with darker red countries like the USA and Australia being more English-dominant. If the web corpus and the Twitter baseline corpus represent similar populations, then the percent of English per country should be highly correlated. By looking at correlations across languages, we can estimate which populations are best represented by the corpus.

**Figure 3. Countries by Percent English Use, CGLU v.4.2**

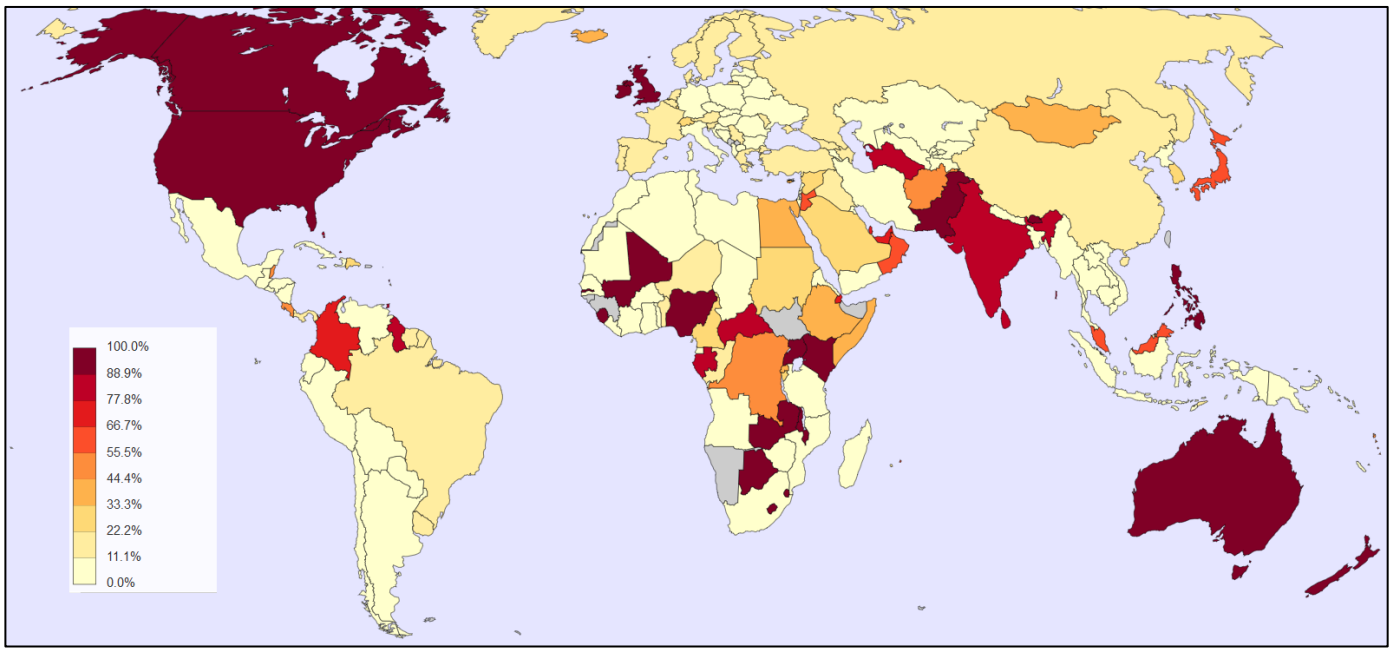

Table 4 shows eight major languages by their total number of words in CGLU v.4.2 and by the correlation on a country-by-country basis between the percent of data in that language with the Twitter baseline corpus. The percent of a language's use in a given country on Twitter is not a ground-truth baseline so much as a point of comparison of the similarity between these sources of digital language use. We notice, first, a quick drop-off in the amount of data per language: English, Spanish, French, and Russian together account for over half of the overall data. Second, the correspondence between different data sources varies widely by language. Vietnamese, which is dominant in Vietnam and some other countries in Southeast Asia but is not found on a global scale, has a similar profile in both datasets. But colonial languages like English and French are used to a different degree across countries. In both cases, web data has more relative use of the colonial language than Twitter data (i.e., English is more widely used on the web than on Twitter). On the other hand, though, languages in the middle of these extremes (e.g., Arabic and Russian), which are used in many countries but are still regional languages, have generally high correlations in their relative usage per country across data sources. This implies that only a few languages like English are disproportionately over-represented in the web corpus.

*Table 4. Comparison of Language Profiles by Country Against Twitter*

|  | N. Words (CGLU v.4.2) | Correlation (Twitter + CGLU v.4.2) | Correlation (Twitter + CGLU v.3) |
|---|---|---|---|
| English | 129.3 billion | 0.686 | 0.738 |
| Spanish | 38.7 billion | 0.907 | 0.935 |
| French | 26.2 billion | 0.692 | 0.648 |
| Russian | 25.4 billion | 0.823 | 0.798 |
| Chinese | 24.5 billion | 0.495 | 0.719 |
| Vietnamese | 16.0 billion | 0.998 | 0.997 |
| Portuguese | 6.2 billion | 0.736 | 0.912 |
| Arabic | 1.3 billion | 0.867 | 0.886 |

The point of this section has been to evaluate these three corpora against one another and against ground-truth demographic data. We know that digital sources of language data are biased towards certain places and populations; much of this bias can be explained by the combination of actual population size together with per capita GDP and estimated internet access. A second important factor is that international languages of communication (English, French, Chinese) are more over-represented in the web corpus. These comparisons help us to understand what populations are being represented in the CGLU and suggest ways in which the corpus could be down-sampled to remove these biases for specific applications.

## 6  Comparing Web Corpora with Twitter Corpora

Previous work has shown that there is a correspondence between (i) digital sources of language data like web corpora (Cook & Brinton, 2017) and Twitter corpora (Grieve, et al., 2019) and (ii) local language use as collected via non-digital sources. Such studies have focused on inner-circle varieties of English (Canada, the US, the UK), in part because these countries represent most existing survey-based and interview-based dialect collection projects. We cannot compare the CGLU v.4.2 against ground-truth language data across all countries and all languages in this same way because such ground-truth data does not exist. Instead we systematically compare the similarity of the web corpus and the Twitter corpus across every language-country pair using standard corpus similarity measures (Kilgarriff, 2001; Fothergill, et al., 2016). Given register variation, we expect some degree of divergence between the two corpora. But, using inner-circle varieties of English as a baseline, this shows where there is a greater divergence between the corpora than expected.

We consider each language-country sub-corpus that contains at least 1 million words in both datasets. This gives us 272 observations distributed across 44 languages, as shown in Table 5. Each of these language-country sub-corpora provides an observation of the similarity between the web and Twitter data. Rather than relying on the $\chi^2$ similarity measure, which is sensitive to differences in overall corpus size, we use the Spearman correlation between unigram frequencies, which is somewhat less accurate than the $\chi^2$ measure in balanced situations but less sensitive to corpus size (Kilgarriff & Rose, 1998). This measure is used to determine the relative distance between the web and Twitter data for each observation. In order to have an adequate comparison across disparate corpus sizes, a frequency threshold of 5 occurrences per 10 million words is applied. The basic idea is that the more similar two sub-corpora are, the more their word frequencies will be ranked in the same order. Only aligned words (above the frequency threshold in both datasets) are relevant to this question. For reference, the complete unigram frequency lists by country and language are available as part of the corpus distribution.[9]

*Table 5. N-Gram Comparison Inventory by Language*

| Language | N. Countries | Language | N. Countries | Language | N. Countries |
| --- | --- | --- | --- | --- | --- |
| Arabic | 20 | Greek | 1 | Romanian | 2 |
| Albanian | 2 | Hindi | 2 | Russian | 8 |
| Azerbaijani | 1 | Hungarian | 1 | Serbo-Croatian | 4 |
| Bengali | 1 | Indonesian | 3 | Sinhala | 1 |
| Bulgarian | 1 | Italian | 4 | Slovenian | 1 |
| Catalan | 2 | Japanese | 1 | Spanish | 37 |
| Czech | 1 | Korean | 1 | Swedish | 2 |
| Danish | 2 | Latvian | 1 | Tagalog | 2 |
| Dutch | 3 | Lithuanian | 1 | Tamil | 2 |
| English | 98 | Macedonian | 1 | Telugu | 1 |
| Estonian | 1 | Marathi | 1 | Turkish | 7 |
| Farsi | 2 | Mongolian | 1 | Ukrainian | 1 |
| Finnish | 1 | Norwegian | 2 | Urdu | 2 |
| French | 20 | Polish | 2 | Vietnamese | 1 |
| German | 9 | Portuguese | 15 | | |

---

[9] https://www.earthlings.io/ngram_download.html

The similarity comparison is organized around both languages and countries: First, which languages are the most similar across registers, with inner-circle Englishes as a baseline? Second, which countries are the most similar across registers and is register-similarity related to demographic variables? We start, in Table 6, with the corpus similarity results for inner-circle Englishes, together with the number of unigrams which have passed the frequency threshold in both the web and Twitter datasets. Australia is an outlier, with a significantly lower similarity and a lower number of shared unigrams. We take the average corpus similarity across these inner-circle varieties, however, as our benchmark for expected register variation across web and Twitter data.

*Table 6. Corpus Similarity (CGLU v.4.2 and Twitter) for Inner-Circle Varieties of English*

| Country | N. Unigrams | Spearman Similarity |
|---|---|---|
| Australia | 21,970 | 0.513 |
| Canada | 32,482 | 0.775 |
| Ireland | 30,741 | 0.761 |
| New Zealand | 30,732 | 0.752 |
| United Kingdom | 23,809 | 0.636 |
| United States | 29,517 | 0.731 |
| **Average** | **28,208** | **0.694** |

Because previous studies of the correspondence between digital data and traditional language samples have focused on English, we use the correlation in Table 6 as our baseline when looking at other languages. Table 7 shows the corpus similarity between datasets by language for the 24 languages with at least two countries in each dataset. The main reasoning here is that, for inner-circle varieties of English, previous work has shown that Twitter data (the main focus) presents similar linguistic patterns as traditional data sources. Our baseline correspondence is a Spearman rank correlation of about 0.700. Languages which fall below this baseline indicate a possible divergence between the populations producing web data and those producing Twitter data. This is precisely what we see for many languages: for example, Arabic (0.469) and Portuguese (0.491) and Tagalog (0.434) are especially low. Overall, there are 32 language-country pairs that are above 0.700; and 20 of these pairs are English. On the one hand, if there was the same amount of variation across registers for all languages, then we could assume that all languages had the same correspondence between digital and non-digital language use that is found in inner-circle Englishes. But the similarity between web corpora and Twitter corpora is consistently lower for languages other than English. Why?

*Table 7. Average Twitter-CGLU(v.4.2) Similarity Across Languages*

| Language | N. Countries | Avg. Similarity | Std. Deviation |
|---|---|---|---|
| ara (Arabic) | 20 | 0.469 | 0.072 |
| cat (Catalan) | 2 | 0.674 | 0.061 |
| dan (Danish) | 2 | 0.620 | 0.083 |
| deu (German) | 9 | 0.598 | 0.068 |
| eng (English) | 98 | 0.634 | 0.082 |
| fas (Farsi) | 2 | 0.644 | 0.058 |
| fra (French) | 20 | 0.579 | 0.066 |

| Language | N. Countries | Avg. Similarity | Std. Deviation |
|---|---|---|---|
| hbs (Serbo-Croatian) | 4 | 0.612 | 0.069 |
| hin (Hindi) | 2 | 0.597 | 0.125 |
| ind (Indonesian) | 3 | 0.607 | 0.102 |
| ita (Italian) | 4 | 0.584 | 0.089 |
| nld (Dutch) | 3 | 0.622 | 0.121 |
| nor (Norwegian) | 2 | 0.557 | 0.141 |
| pol (Polish) | 2 | 0.525 | 0.083 |
| por (Portuguese) | 15 | 0.491 | 0.060 |
| ron (Romanian) | 2 | 0.655 | 0.002 |
| rus (Russian) | 8 | 0.584 | 0.033 |
| spa (Spanish) | 37 | 0.565 | 0.088 |
| sqi (Albanian) | 2 | 0.687 | 0.034 |
| swe (Swedish) | 2 | 0.642 | 0.059 |
| tam (Tamil) | 2 | 0.741 | 0.001 |
| tgl (Tagalog) | 2 | 0.435 | 0.065 |
| tur (Turkish) | 7 | 0.471 | 0.045 |
| urd (Urdu) | 2 | 0.653 | 0.026 |

To answer this, we first look at the similarity between country-language sub-corpora with a focus on countries: is there a geographic source of lower similarity that transcends languages? There are 80 countries with at least two languages in both datasets. There is a general geographic effect, with countries ranging from above 0.70 (Sri Lanka, Mexico, Indonesia) to below 0.45 (Brazil, Argentina). But these geographic effects are not highly related to demographic variables like percent of internet usage (r = 0.15) and per capita GDP (r = 0.10). The answer seems to be, then, that variations in the similarity between web corpora and Twitter corpora are organized around language and not related to the population demographics of specific countries.

Second, in order to understand the causes of variation in the similarity between these two data sources, we look at reciprocal similarity relationships between countries within each register. In other words, we look at all countries with an Arabic web corpus and compare the similarity of each country's corpus with every other country's corpus. The average of these similarity values represents the relative degree of difference for each language in each register. The lower the value, the more a data source varies within itself by country. For example, Danish is very consistent across countries in the web corpus (0.812) but Norwegian is quite different across countries (0.475).

*Table 8. Variation By Country Within Data Sources*

| Language | Similarity (CC) | Std. Dev (CC) | Similarity (TW) | Std. Dev (TW) |
|---|---|---|---|---|
| ara (Arabic) | 0.628 | 0.166 | 0.572 | 0.086 |
| cat (Catalan) | 0.741 | 0.000 | 0.700 | 0.000 |
| dan (Danish) | 0.812 | 0.000 | 0.704 | 0.000 |
| deu (German) | 0.732 | 0.039 | 0.690 | 0.064 |
| eng (English) | 0.758 | 0.042 | 0.667 | 0.089 |
| fas (Farsi) | 0.687 | 0.000 | 0.657 | 0.000 |

| Language | Similarity (CC) | Std. Dev (CC) | Similarity (TW) | Std. Dev (TW) |
|---|---|---|---|---|
| fra (French) | 0.731 | 0.047 | 0.635 | 0.082 |
| hbs (Serbo-Croatian) | 0.672 | 0.077 | 0.649 | 0.075 |
| hin (Hindi) | 0.474 | 0.000 | 0.322 | 0.000 |
| ind (Indonesian) | 0.520 | 0.098 | 0.580 | 0.058 |
| ita (Italian) | 0.792 | 0.035 | 0.691 | 0.074 |
| nld (Dutch) | 0.764 | 0.041 | 0.635 | 0.142 |
| nor (Norwegian) | 0.475 | 0.000 | 0.400 | 0.000 |
| pol (Polish) | 0.797 | 0.000 | 0.653 | 0.000 |
| por (Portuguese) | 0.726 | 0.043 | 0.601 | 0.077 |
| ron (Romanian) | 0.743 | 0.000 | 0.649 | 0.000 |
| rus (Russian) | 0.736 | 0.054 | 0.770 | 0.049 |
| spa (Spanish) | 0.735 | 0.048 | 0.622 | 0.104 |
| sqi (Albanian) | 0.661 | 0.000 | 0.764 | 0.000 |
| swe (Swedish) | 0.802 | 0.000 | 0.671 | 0.000 |
| tam (Tamil) | 0.802 | 0.000 | 0.739 | 0.000 |
| tgl (Tagalog) | 0.550 | 0.000 | 0.262 | 0.000 |
| tur (Turkish) | 0.693 | 0.043 | 0.512 | 0.050 |
| urd (Urdu) | 0.839 | 0.000 | 0.703 | 0.000 |
| **AVG** | **0.703** | **0.031** | **0.619** | **0.040** |

The similarity measures in Table 8 show that the lower correspondence between the data sources outside of inner-circle Englishes comes from the generally lower similarity within the CGLU v.4.2 corpus (an average of 0.619 as opposed to 0.703 for Twitter). While there is a strong Pearson correlation between similarity measures across languages (0.776), the Twitter data set in general has more similar corpora across countries than CGLU v.4.2. One reason, of course, is that Twitter constitutes a single register while web corpora encompass several sub-genres. But the influence of different registers within the web corpus is hard to quantify here. Another potential reason is that the CGLU is significantly larger than the Twitter baseline corpus (423 billion words vs 8 billion). However, there is not a significant correlation between corpus size per language and the average similarity values per language. Thus, the sheer size of the web corpus is not the cause. Another potential reason is that the web corpus represents a more diverse population from each country. This is hard to quantify, although we can look at the diversity of language use within each corpus. The top twenty languages account for 91.1% of the web data and 90.5% of the Twitter data. In a non-geographic sense, then, both data sets have a similar degree of linguistic diversity.

This section has used standard corpus similarity measures to investigate relationships between sub-corpora of the web and Twitter data organized by country and by language. First, we have seen that the most proto-typical digital language data (English from inner-circle countries) has some of the highest similarities across data sources. The similarity between data sources varies across both languages and countries, but it is not correlated with demographic attributes of countries or with the relative amount of data per language. Within data sources, there is higher similarity within languages across countries in Twitter data than in web data, although the two are highly correlated on this measure. The purpose of this evaluation is to systematically represent relationships within and between corpora, an especially important task because most work connecting digital data to specific local language use has used only inner-circle varieties of English.

## 7   Data Structure and License

The full version of the web corpus is available under the GNU-GPL v.3 License[10], including ngram frequency lists for the web data and the Twitter data by language and country.[11] The corpus itself is stored in compressed csv files with the following columns: Language, URL, Number of Words, Text. Each web page is a single row and paragraph breaks are retained within samples as line breaks. Each file is limited to 100k web pages. The corpus is organized by folder: Region > Country > Language. The ngram dataset with unigram frequencies is organized by register (web, Twitter), with folders for each language. The code for creating the corpus is available[12], as well as the code for the language identification software.[13] Additionally, an interactive visualization component is available for further exploring these datasets.[14]

## 8   Conclusions

The main contribution of this paper is to systematically evaluate the relationship between digital sources of language data (the web and Twitter) against one another and against population demographic data on a global scale across many languages. The distribution of both datasets is best explained by a combination of (i) country-level population density, (ii) relative access to internet technologies, and (iii) per capita GDP. While there remains variation to be explained, these three factors explain much of the corpus distribution. Delving more deeply into corpus similarity measures, the paper has shown that these two sources of digital data agree most when representing inner-circle varieties of English, precisely those contexts which have been the focus of previous work on validating digital datasets. Variation within and between both datasets is structured more around individual languages and is less predictable given country-specific population and corpus size information. Work based on previous versions of the corpus (Dunn, 2019a, 2019b) have shown that meaningful dialectal variation can be modeled using this source of data. The internal (corpus similarity) and external (demographic) evaluations in this paper strongly suggest that future work based on these expanded country-language sub-corpora will support further advances in corpus-based dialectology.

The secondary contribution of this paper is to describe a publicly-available corpus that greatly expands upon currently available geo-referenced text data. This dataset provides gigaword corpora for 31 languages and 59 countries. Importantly, this is made possible by a language identification model that maintains high accuracy across many languages in a multi-register, short-text experimental paradigm. This is important for working with less-commonly used languages in large web-crawled datasets, a problem that is growing as digital language use becomes a primary means of communication.

---

[10] https://www.earthlings.io/corpus_download.html

[11] https://www.earthlings.io/ngram_download.html

[12] https://github.com/jonathandunn/common_crawl_corpus

[13] https://github.com/jonathandunn/idNet

[14] https://www.earthlings.io and https://github.com/jonathandunn/earthlings